\documentclass{article}
\usepackage{ijcai19}
\usepackage{float}
\usepackage{times}
\usepackage[utf8]{inputenc}
\usepackage[small]{caption}
\usepackage{graphicx}
\usepackage{amsmath}
\usepackage{mathtools}
\usepackage{amssymb}
\usepackage{booktabs}
\usepackage{url}
\newcommand{\IGNORE}[1]{}
\makeatletter
\newcases{lrdcases}
{\quad}
{$\m@th\displaystyle{##}$\hfil}
{$\m@th\displaystyle{##}$\hfil}
{\lbrace}
{} 

\newcases{lrdcases*}
{\quad}
{$\m@th\displaystyle{##}$\hfil}
{{##}\hfil}
{\lbrace}
{} 
\makeatother
\newcommand{\tl}{\raisebox{-.6ex}{\symbol{'176}}}

\title{Swarm Engineering Through Quantitative Measurement of Swarm Robotic Principles in a 10,000 Robot Swarm}

\author{
  John Harwell\footnote{Contact Author}\And
  Maria Gini\\
  \affiliations
  Department of Computer Science and Engineering, University of Minnesota\\
  \emails
  \{harwe006,gini\}@umn.edu
}

\begin{document}

\maketitle

\begin{abstract}
  When designing swarm-robotic systems, systematic comparison of algorithms from
  different domains is necessary to determine which is capable of scaling up to
  handle the target problem size and target operating conditions. We propose a set of
  quantitative metrics for scalability, flexibility, and emergence which are capable
  of addressing these needs during the system design process. We demonstrate the
  applicability of our proposed metrics as a design tool by solving a large object
  gathering problem in temporally varying operating conditions using iterative
  hypothesis evaluation. We provide experimental results obtained in simulation for
  swarms of over 10,000 robots.
\end{abstract}

\section{Introduction}\label{sec:intro}

Swarm Robotics (SR) systems consist of large numbers of relatively few homogeneous
groups of robots \cite{Sahin2005}. The main differentiating factors between SR
systems and multi-agent robotics systems stem from the origins of SR as an offshoot of
Swarm Intelligence (SI), which investigates algorithms and problem solving techniques
inspired from natural systems such as bees, ants, and termites \cite{Labella}. The
main properties are:

\emph{Scalability}. SR systems have no centralized control/single point of failure,
which makes them scalable to hundreds or thousands of agents, much like their natural
counterparts \cite{Lerman}.

\emph{Emergence}. Agents in SR systems collectively find solutions to a problem at
hand that they cannot solve alone \cite{Cotsaftis2009}. More precisely, emergent
behaviors arising through this collective search process refer to the appearance of
self-organization within the system due to robot
interactions~\cite{Winfield2005a,Galstyan2005}. Such behaviors cannot be predicted
from individual agent behaviors (i.e., it is the set difference between micro and
macro swarm behavior \cite{Szabo2014}).

\emph{Flexibility}. In SR systems all decisions are made by individual agents based
on locally available information from neighbors as well as their own limited sensor
data. This results in reactive and adaptive swarm-level behaviors that attempt to
mitigate adversity and exploit beneficial changes in dynamic environmental
conditions and/or problem definitions \cite{Harwell2018,Winfield2008}.

\emph{Robustness}. The number of agents in a SR system is unlikely to remain constant
during execution, and can fluctuate due to introduction of new agents and robotic
failures. A high failure rate within a swarm may slow its operation, but it does not
prevent the accomplishment of its objective \cite{Lerman}, in contrast to many
multi-agent systems which cannot withstand such losses.

The duality between SR and natural systems enables effective parallels to be drawn
with many naturally occurring problems, such as foraging, collective transport of
heavy objects, environmental monitoring/cleanup, self-assembly, exploration, and
collective decision making~\cite{Sahin2005,Hecker2015}. As a result, SR systems are
especially well suited to tackle complex tasks in environments where robustness and
flexibility of robotic systems is key to success, such as space
\cite{Rouff2004}. Other real-world problems with applicable SR solutions include
tracking lake health, clearing a corridor on a mining operation, hazardous material
cleanup, and search and rescue~\cite{Sahin2005,Hecker2015,Labella}.

In a foraging task, robots gather objects from across a finite operating arena and
bring them to a central location under various conditions and constraints. Foraging
is one of the most extensively studied applications of SR, due to its straightforward
mapping to real-world applications~\cite{Hecker2015}. In this paper, we study
scalability, emergence, and flexibility in the context of a foraging task.

We present measurement methods of swarm scalability and emergent self-organization
using projected vs.~observed performance increases and expected vs.~actual
performance losses, respectively, as swarm sizes increase. We show that they provide
valuable input into iterative SR system design and algorithm selection for a
particular problem/swarm size. We also present a methodology to measure swarm
flexibility (its ability to react and adapt to changing environmental conditions)
using mathematical methods of curve similarity. By correlating temporal curves of
swarm performance with curves of varying environmental conditions, we can quantify
how much performance should be expected to be gained or lost between scenarios with
different operating conditions or task parameters. We show that our methodology
enables mathematically grounded algorithm selection decisions through quantitative
comparisons of similarity between the target operating conditions and those in which
the algorithm was developed and tested.

We evaluate our proposed metrics using a simulated design process to demonstrate the
correctness of predictive hypotheses framed using our measures as swarm sizes are
scaled up to 16,384 robots. Results indicate that explicit quantification of
different facets of a swarm's intelligence across all tested scales is possible using
our proposed measures. This quantification is shown to provide insight into what
properties the ``best'' method for solving the problem at hand should have by
exposing predictive differences in swarm response to external and internal stimuli
not evident from raw performance curves.

\subsection{Background and Related Work}\label{ssec:bg-and-rw}

In recent years, many theoretical SR system design tools have become available
\cite{Matthey2009,Correll2008,Lopes2016a,Hamann2013}. These tools have made it easier
to conduct mathematical analysis of algorithms and derive analytical, rather than
weakly inductive proofs of correctness~\cite{Winfield2008}. Despite this, there has
not been a corresponding increase in the average swarm sizes used to evaluate new
algorithms (Notable exceptions include~\cite{Lopes2016a} (600
robots),~\cite{Hecker2015} (768 robots),~\cite{Hamann2008} (375 robots)).
Investigation of simple behaviors such as pattern formation, localization, or
collective motion, where design and computational complexity do not inherently limit
scalability, is generally evaluated with relatively small scales (\tl40 robots
\cite{Winfield2008}). Methods for more complex behaviors, such as
foraging~\cite{Ferrante2015,Pini2011a} (20 robots),~\cite{Pini2011b} (30 robots),
task allocation~\cite{Correll2008} (25 robots) are likewise tested at similar scales.

We believe that the lack of evaluation of large SR systems is partially due to a lack
of standardized comparison methods to accurately characterize swarm performance
across problem scales and domains. Building on previous works
\cite{Harwell2018,Hecker2015}, we propose a swarm scalability measure analogous to
projected vs.~observed speedup when core count is increased on supercomputing
clusters.

Within SR, no widely accepted theory of self-organizing systems
exists~\cite{Hamann2008,Galstyan2005}. Cotsaftis \shortcite{Cotsaftis2009} presents a
control-theoretic model of emergence, distinguishing between \emph{complicated}
systems which can be studied by the methods of scientific reductionism, and
\emph{complex} systems which originate from the existence of a threshold above which
interaction between system components overtakes outside interactions, leading to
system self-organization and new behavior not predictable from component study,
similar to~\cite{George2005}. While many papers cite evidence that their algorithms
exhibit emergent behavior \cite{Liu2007a,Frison2010,Harwell2018,Matthey2009}, or even
prove simple emergent properties via temporal logic (\cite{Winfield2005a}), few
provide a quantitative method for measuring emergence (with the exception
of~\cite{Szabo2014}, who used robot nearest-neighbor calculations to calculate a
degree of interaction for the swarm). We present an empirical method for measuring
the level of self organization present in a swarm by measuring the linearity of
inter-robot interference as the swarm size is increased, as a small step towards the
development of a more general emergence theory.

While it is common to evaluate swarms under conditions different than their
development, the resultant claims of flexibility due to performance similarity across
scenarios is strictly qualitative, due to a lack of mathematical quantification of
``difference'' in scenario conditions as a factor in performance comparisons. We
utilize curve similarity methods from Jekel \emph{et al.}~\shortcite{Jekel2018}, who tested
different mathematical curve similarity measures in various noise and normalization
conditions to provide the required mathematical basis. We derive a difference measure
between expected and observed performance in temporally variable environmental
conditions, and use it to quantify the swarm \emph{reactivity} (how closely swarm
performance tracks changing environmental conditions in time) and \emph{adaptability}
(how well the swarm exploits or resists beneficial or adverse changes in
environmental conditions).

\section{Proposed Quantitative Measurements}\label{sec:measurements}
\subsection{Swarm Scalability}\label{ssec:meas-scalability}

Hecker and Moses \shortcite{Hecker2015} calculated scalability $S(N)=P(N)/N$ as per-robot
efficiency using a performance measure $P(N)$, where $N$ is the size of the
swarm. $P(N)$ is general in nature, and can measure time to complete a certain number
of tasks, task completion rate, etc. While this measure provides some insight into
scalability, it is not predictive (e.g.~given $P(N)$, we cannot plausibly estimate
$P({2}{N})$ without retroactively charting $S(N)$ across a range of $N$).

We extend $P(N)$ to $P(N,\kappa,t)$, where $\kappa$ is the swarm control
algorithm plus algorithmic parameters and $t$ is the current discrete timestep (swarm
performance varies temporally, tracing out a performance curve). For a simulation of
length $T$ broken into $t$ timesteps, our generalized formulation allows us to
simultaneously analyze (1) cumulative performance by summing across all $t\in{T}$,
(2) temporally varying performance curves via pairwise point comparison for each
$t\in{T}$ (see Section~\ref{ssec:meas-flexibility}).

We define a more predictive scalability metric for use as an iterative system design
tool using the Karp-Flatt metric~\cite{Karp1990}. Traditionally, the
\emph{serial fraction} $\mathbf{e}$ of the Karp-Flatt metric measures the level of
parallelization of a particular program, with smaller values indicating high
parallelization and plausibly expected speedups if more computational resources are
added. In SR, it measures the intelligence of a swarm by exposing the part of
$P(N,\kappa,T)$ that did not utilize inter-robot cooperation.

\smallskip
\noindent {\bf Scalability: } $\mathbf{e}(N_1,N_2,\kappa)$ for two swarms of sizes
$N_1$ and $N_2$ ($N_2 > N_1$) controlled by a method $\kappa$ is defined as:
\begin{equation}
  \label{eqn:karp-flatt}
  \mathbf{e}(N_1,N_2,\kappa) = \frac{\frac{1}{\phi(N_1,N_2,\kappa)} - \frac{1}{N_1}}{1- \frac{1}{N_1}}
\end{equation}
where
\begin{equation}
  \label{eqn:projective-performance}
  \phi(N_1,N_2,\kappa) = \sum_{t\in{T}}\frac{P(N_2,\kappa,t)}{\frac{N_2}{N_1}~P(N_1,\kappa,t)}
\end{equation}
is based on a swarm's projected vs.~observed performance.  Intuitively,
Eqn.~\eqref{eqn:projective-performance} can be understood in terms of computational
workloads: if a job takes $T$ seconds with $N$ resources, ideally it will take $T/2$
seconds with $2N$ resources in a perfectly parallelizable system.

\subsection{Swarm Emergence via Self-Organization}\label{ssec:meas-self-org}

We can infer from Cotsaftis \shortcite{Cotsaftis2009} that SR systems with a high degree of local
interactivity should exhibit higher levels of self-organization (and therefore
emergent behavior, as established by \cite{George2005}) than systems with a low
degree of interactivity~\cite{Szabo2014,Galstyan2005,Winfield2005a}. We can therefore
approximately measure a swarm's emergent behavior by measuring its self organization.

We calculate (post-hoc) the expected number of robots engaged in collision avoidance
(as opposed to doing useful work) in a swarm of size $N$, and use this to derive the
amount of time lost due to inter-robot interference on each timestep $t$, denoted as
$t_{lost}^{N}(t)$. This can then be used to compute the per-timestep fraction of
overall performance loss ($P_{lost}(N,\kappa,t)$) due to inter-robot interference as
follows:
\begin{equation}\label{eqn:fl-plost}
  P_{lost}(N,\kappa,t) =
  \begin{cases}
    {P(1,\kappa,t)}{t_{lost}^{1}(t)} & \text{if N = 1}
    \\
    P(N,\kappa,t){t_{lost}^{N}(t)} \\ ~ - {N}{P_{lost}(1,\kappa,t)}& \text{if N  $>$ 1}
    \\
  \end{cases}
\end{equation}
For $N \ge 1$ we subtracted the interference that would have occurred in a
non-interactive swarm of $N$ robots (i.e.~a swarm of size $N$ which only interacted
with arena boundaries). Let $M=\{1,2,4,\ldots{},m_{max}$\} be a logarithmically
distributed (powers of two) set. If we then compute Eqn.~\eqref{eqn:fl-plost} for
each $m\in{M}$, sub-linear fractional losses between $P_{lost}(m_{i-1},\kappa,t)$ and
$P_{lost}(m_i,\kappa,pt)$ indicate that the method $\kappa$ is scalable in the
neighborhood of sizes near $m_i$ (i.e., doubling the swarm size does not double the
amount of inter-robot interference).

Using Eqn.~\eqref{eqn:fl-plost}, we quantify a swarm's ability to detect and
eliminate non-cooperative situations between agents~\cite{George2005} (e.g., frequent
inter-robot interference) as swarm size is increased. Our self-organization measure
can therefore be used to measure the ``intelligence'' of different algorithms
deployed in a swarm in terms of their to manage space without sacrificing
performance.

\smallskip
\noindent {\bf Self-organization:} $Z(m_i,\kappa)$ is defined as:
\begin{equation}\label{eqn:self-org}
  Z(m_i,\kappa) = \sum_{t\in{T}}1 - \frac{1}{1 + e^{-\theta_Z(m_i,\kappa,t)}}
\end{equation}
where
\begin{equation}\label{eqn:self-org-t-theta}
  \theta_Z(m_i,\kappa,t) = P_{lost}(m_i,\kappa,t) - \frac{m_i}{m_{i-1}}{P_{lost}(m_{i-1},\kappa,t)}
\end{equation}
Intuitively, Eqn.~\eqref{eqn:self-org-t-theta} indicates that if the proportional
increases in fractional performance losses observed at swarm sizes $m_{i-1}$ and
$m_i$ are sublinear ($\theta_Z(m_i,\kappa,t) \le 0$), then
Eqn.~\eqref{eqn:self-org} $\rightarrow 1.0$, indicating that self-organization
occurred, and vice versa with Eqn.~\eqref{eqn:self-org} $\rightarrow 0.0$ for
superlinear increases. We note that this definition of self-organization assumes that
a swarm is operating in a bounded space; this is not generally a limiting assumption,
as such restrictions are common in real-world applications such as
warehouses~\cite{Pini2011b}. 

\subsection{Swarm Flexibility via Reactivity and Adaptability}\label{ssec:meas-flexibility}

We define a swarm's \emph{flexibility} in terms of its ability to adapt to changes in
the external environment over time, which can include (1) costs of performing a
particular action (e.g., picking up/dropping an object in the case of foraging), (2)
maximum robot speed, modeling changing environmental conditions (e.g., wheeled robots
buffeted in variable winds) or the performance of some tasks more slowly than others
(e.g., carrying objects of different sizes during foraging).

Let $I_{ec}(t)$ and $V_{dev}(t)$ be continuous, one-dimensional signals representing
respectively (1) the ideal environmental conditions that a given swarm was developed
in, (2) the waveform of a deviation from $I_{ec}(t)$. We have restricted our study to
one-dimensional characterizations of environmental conditions, but extensions to
higher dimensions should be relatively straightforward.

Intuitively, we define swarm \textit{reactivity} $R(N,\kappa)$ as how closely the
observed performance curve $P(N,\kappa,t)$ tracks an applied variance
$V_{dev}(t)$. In a swarm with \textit{optimal reactivity} $R^*(N,\kappa)$,
$P(N,\kappa,t) = c_t{V_{dev}(t)}$, where $c_t$ is a non-negative per-timestep
constant (i.e., instantaneous tracking).

\smallskip
\noindent {\bf Reactivity $R(N,\kappa)$} is defined as:
\begin{equation}\label{eqn:reactivity}
  R(N,\kappa) = DTW(P_{R^{*}}(N,\kappa,t),P(N,\kappa,t))
\end{equation}
where we define $DTW(X,Y)$ to be a \textit{Dynamic Time Warp} similarity measure between
two discrete curves $X$ and $Y$, based upon the conclusions in~\cite{Jekel2018}. We
note that (1) DTW correctly matches temporally shifted sequences of applied variance
and observed performance (achieving $R^*(N,\kappa)$ is not possible for realistic
swarms), (2) DTW exhibits robustness to signal noise, and SR systems are inherently
stochastic.

Formally, we derive $R(N,\kappa)$ for a stepped experiment of length $T$ using
$Q : [X,a,b] \rightarrow X'$, with $a,b \in \mathbb{N}$, as a mapped min-max
normalization of a curve $X$ into the range $[a,b]$:
\begin{equation}
  X' = \frac{(b - a)(X - \min{X})}{\max{X} - \min{X}} + a
\end{equation}
We construct the optimal performance curve $P_{R^*}(N,\kappa,t)$ of a maximally
reactive method $\kappa$ by normalizing the difference curve of
$V_{dev}(t) - I_{ec}(t)$ to the scale of the observed performance under ideal
conditions:
\begin{equation}\label{eqn:reactivity-star}
\begin{aligned}
  P_{R^{*}}(N,\kappa,t) = Q(&V_{dev}(t),\\ &\min(P_{ideal}(N,\kappa,t)), \\
  &\max(P_{ideal}(N,\kappa,t)))
  \end{aligned}
\end{equation}
Finally, the \textit{adaptability} $A(N,\kappa)$ of a swarm can be thought of
as its ability to (1) minimize performance losses under adverse conditions
($V_{ec}(t) \ge I_{ec}(t)$), (2) proportionally exploit beneficial deviations from
ideal conditions ($V_{ec}(t) < I_{ec}(t)$).

\smallskip
\noindent {\bf Adaptability:} $A(N,\kappa)$ is defined as the $DTW(X,Y)$ similarity
to the \textit{optimal adaptability} curve $P_{A^{*}}(N,\kappa,t)$:
\begin{equation}\label{eqn:adaptability}
  A(N,\kappa) = DTW(P_{A^{*}}(N,\kappa,t),P(N,\kappa,t))
\end{equation}
Formally, we define $P_{A^{*}}(N,\kappa)$ as:
\begin{equation}\label{eqn:adaptability-star}
  P_{A^{*}}(N,\kappa,t) =
  \begin{cases}
    \frac{{c_t}V_{ec}(t)}{I_{ec}(t)}R(N,\kappa) & \text{if $V_{ec}(t) < I_{ec}(t)$} \\
    P_{ideal}(N,\kappa,t) & \text{else}
  \end{cases}
\end{equation}
where $c_t$ is a positive per-timestep constant.

We briefly note that a weighted combination of
Eqn.~\eqref{eqn:reactivity} and Eqn.~\eqref{eqn:adaptability} could be employed as a design
tool to gain insight into
the most suitable method for a set of target operating conditions, and reduce
the simulation-reality gap~\cite{Hecker2015,Brutschy2015}.

\section{Example of Application to a Foraging Task}\label{exp-application}

We evaluate our proposed methodology by simulating the iterative SR design process
for a large-scale foraging application that (1) has 10,000 robots available for use
(not all robots need to be utilized) (2) has an operating area $\ge 4,000 m^{2}$, (3)
has temporally variable operating conditions which can suddenly change from favorable
to unfavorable (or vice versa) at some point during system operation (i.e., a passing
storm in an outdoor environment). 

We utilize our derived metrics to iteratively form
and test hypotheses as we scale swarm and arena size to the parameters of our target
application in order to refine our estimate of which method $\kappa$ provides maximum
performance.

We use homogeneous swarms and the following candidates controllers
$\kappa\in$\{CRW,DPO,GP-DPO\} from \cite{Harwell2018}, summarized briefly here. In
\textit{Correlated Random Walk (CRW)} swarms, robots perform a correlated random walk
until they acquire an object, which they then transport to the nest using phototaxis
(i.e., motion in response to light). In \textit{Decaying Pheromone Object (DPO)}
swarms, robots track seen objects using exponentially decaying pheromones, and
determine the ``best'' object to acquire using derived information relevance. In
\textit{Greedy Partitioning DPO (GP-DPO)} swarms, robots stochastically choose to do
either (1) the entire foraging task themselves (i.e., retrieving an object and
bringing it to the nest), or (2) one of two subtasks of bringing an acquired object
to an intermediate drop site (cache), or picking one up from a cache and bringing it
to the nest.

\begin{figure}[h]
 \centering
  \includegraphics[width=.95\linewidth]{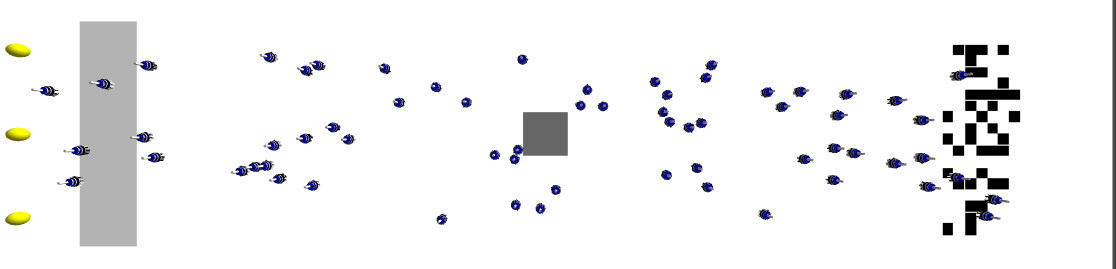}
  \caption{\label{fig:ss-foraging-long} A screen shot of a simulation with multiple
    robots, objects to be collected, and the nest on the left side.  }
 \end{figure}

The experiments described in this paper have been carried out in the ARGoS
\cite{Pinciroli2012} simulator. We employ a dynamical physics model of the robots in
a three dimensional space for maximum fidelity (robots are still restricted to motion
in the XY plane), using a model of the s-bot developed by~\cite{Dorigo2005c}.  We use
a single-source foraging experiment design~\cite{Harwell2018,Ferrante2015,Pini2011a},
in which all objects are concentrated at the far end of a rectangular, obstacle-free
arena, shown in Fig.\ref{fig:ss-foraging-long}. For all experiments we average the
results of 50 experimental runs of $T = 10,000$ seconds for each $\kappa\in$
\{CRW,DPO,GP-DPO\}.

\subsection{Ideal Conditions}\label{ssec:res-ideal-cond}

\begin{figure}[tb]
  \centering
  \includegraphics[width=.9\linewidth]{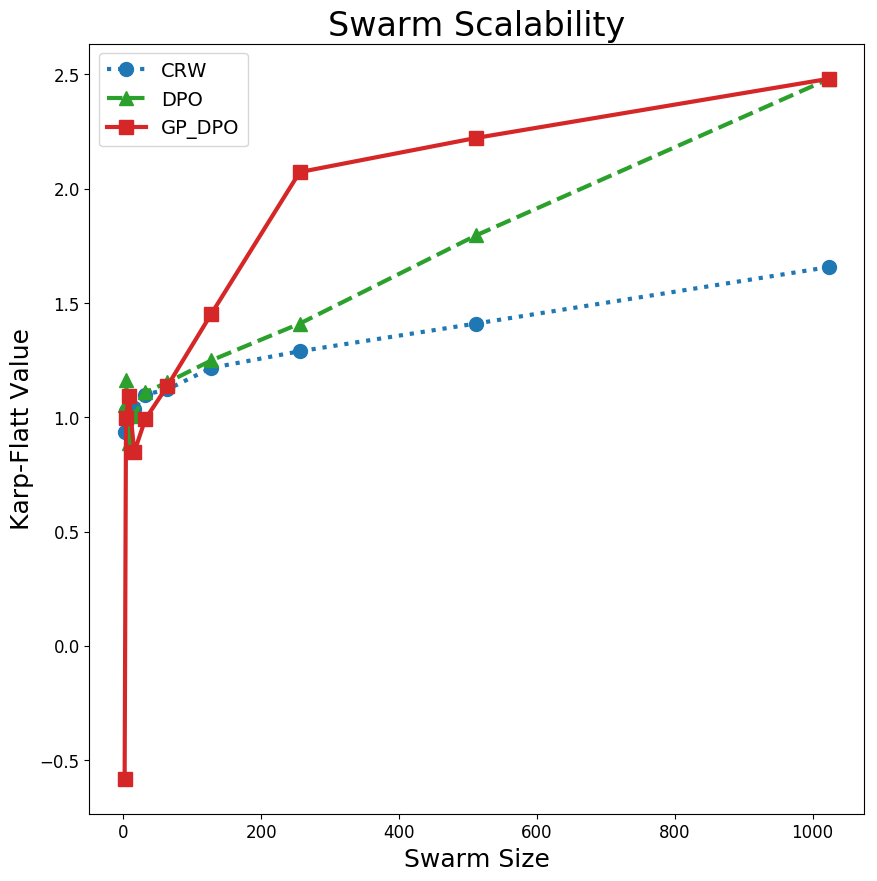}
  \caption{\label{fig:ideal-scalability} Swarm scalability
    $\mathbf{e}(N_1,N_2,\kappa)$ for the 32$\times$16 scenario. CRW swarms are the
    most parallelizable (i.e., proportional performance increases are likely at
    higher values of $N$).}
\end{figure}

We begin with a $32\times16=368~m^2$ arena, and define $P(N,\kappa,T)$ for ideal
operating conditions as the cumulative number of objects gathered within a simulation
up to time $t = T$. We first evaluate our scalability measure on small swarms of up
to 1,024 robots in this arena (Fig.~\ref{fig:ideal-scalability}).

\begin{figure}[tb]
  \centering
  \includegraphics[width=.9\linewidth]{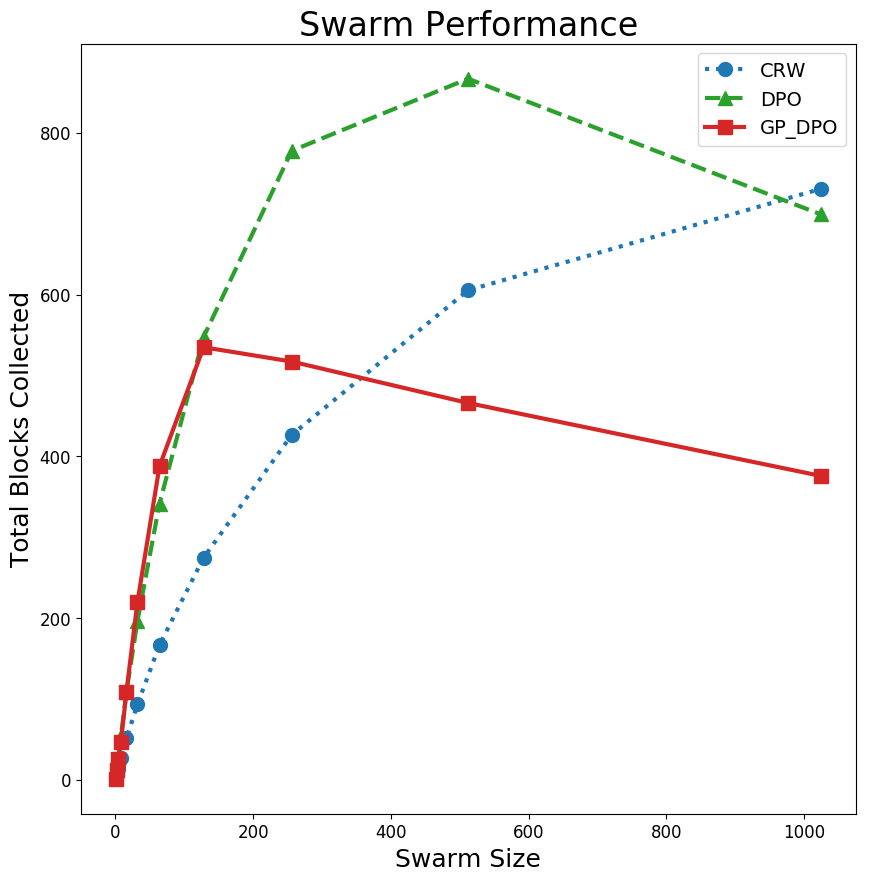}
  \caption{\label{fig:ideal-perf-32x16} Swarm performance $P(N,\kappa)$ for the
    $32\times16$ scenario.}
\end{figure}

The results in Fig.~\ref{fig:ideal-scalability} confirm our intuition that as swarm
sizes begin to approach natural scales in confined, obstacle-free spaces, randomized
motion is the most scalable navigation method. However, looking at the performance
curves for the $32\times16$ scenario in Fig.~\ref{fig:ideal-perf-32x16}, we see that
for smaller swarm sizes, the more ``intelligent'' DPO/GP-DPO approaches perform
better.  The trendlines for scalability in Fig.~\ref{fig:ideal-scalability} and
performance in Fig.~\ref{fig:ideal-perf-32x16} for DPO and GP-DPO swarms are
consistent: the asymptotic behavior of the exponential performance curves (the
characteristic s-shape of swarms) are ordered according to trends of the scalability
curves, demonstrating the predictive insight possible with our metric.

\begin{figure}[tb]
  \centering
  \includegraphics[width=.9\linewidth]{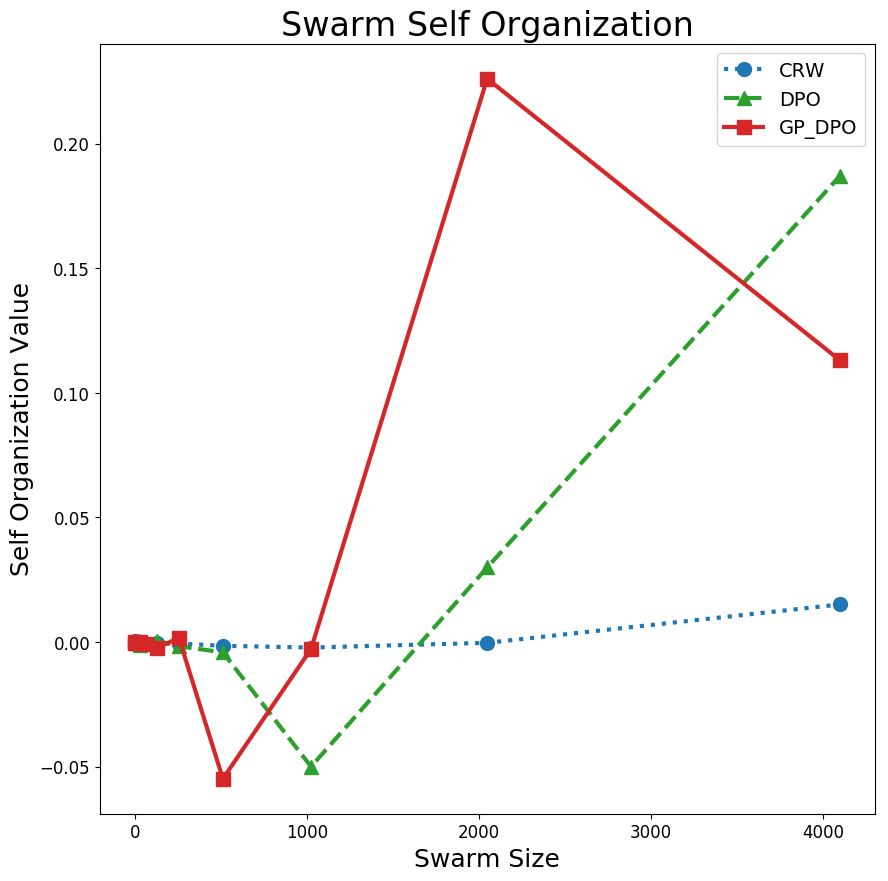}
  \caption{\label{fig:ideal-self-org-64x32} Swarm self-organization $Z(N,\kappa)$ for
    the $64\times32$ scenario. Asymptotic trendlines suggest the hypothesis that DPO
    swarms have the greatest potential for self-organization with $N > 4,096$.}
\end{figure}

\begin{figure}[tb]
  \centering
  \includegraphics[width=.9\linewidth]{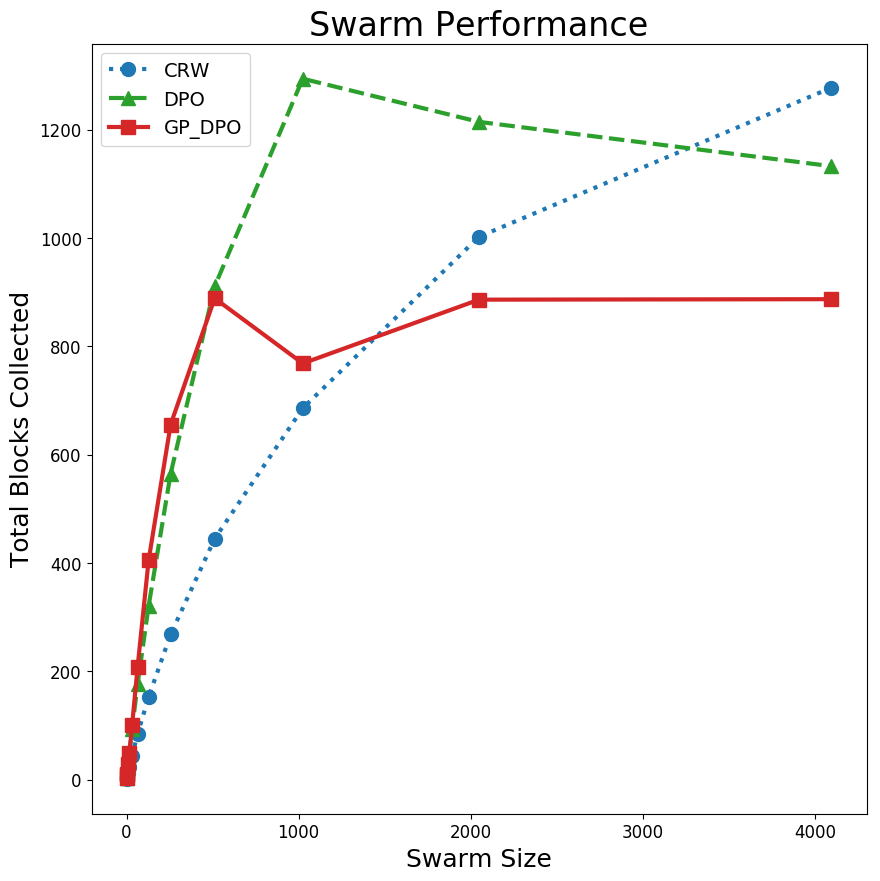}
  \caption{\label{fig:ideal-perf-64x32} Swarm performance $P(N,\kappa)$ for the
    $64\times32$ scenario.}
\end{figure}

Scaling up swarm/problem size to 4,096 robots and $64\times32 = 1,388~m^2$ arena
respectively, we evaluate our self-organization metric
(Fig.~\ref{fig:ideal-self-org-64x32}). Correlating the self-organization and
performance curves in Fig.~\ref{fig:ideal-self-org-64x32} and
Fig.~\ref{fig:ideal-perf-64x32}, respectively, we see that for DPO and GP-DPO swarms
the sizes at which maximum $P(N,\kappa,T)$ and first negative $Z(N,\kappa)$ occur are
approximately identical, demonstrating the usefulness of our self-organization metric
as a predictive design tool (i.e.~no need to continue to scale up a method once you
observe a negative $Z(N,\kappa)$ value). Finally, in CRW swarms we observe that the
consistently low levels of self-organization nevertheless correlate with performance
trendlines suggesting the highest possible performance of the three tested methods at
large sizes. In mathematical terms, we gain the insight that the second
derivative of the self-organization curves is just as important as the first
derivative (which gives the asymptotic trendline of the self-organization for a given
$\kappa$) in predicting asymptotic performance. We hypothesize that this is due to a
threshold problem/swarm size above which the more ``intelligent'' DPO/GP-DPO methods
break down due to high levels of inter-robot interference, while the more simplistic
CRW method is able to more efficiently scale beyond it.

\begin{figure}[tb]
  \centering
  \includegraphics[width=.9\linewidth]{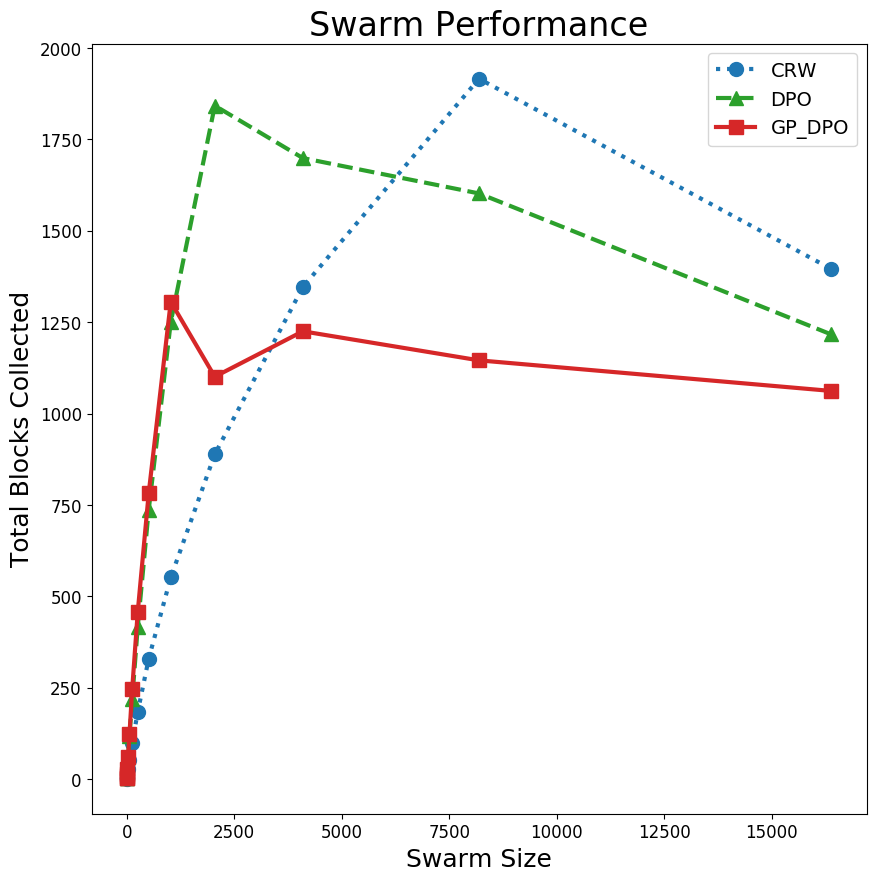}
  \caption{\label{fig:ideal-perf-96x48} Swarm performance $P(N,\kappa)$ across all
    $m \in M$ for the 96$\times$48 scenario.}
\end{figure}

Building on these insights, we hypothesize CRW swarms will have the highest level of
performance as we scale swarm/problem size to 16,384 robots and a
$96\times48 = 4,608~m^2$ arena respectively, which is confirmed by
Fig.~\ref{fig:ideal-perf-96x48}.

\subsection{Temporally Varying Conditions}\label{ssec:res-tv-step}

We next evaluate our three candidate approaches with temporally varying operating
conditions. We apply throttling functions to the maximum robot speed while carrying a
block using the Heaviside $H(t)$ step functions with amplitude
$\beta=\{0.1, 0.2, 0.4, 0.8\}$, and set $\alpha=\frac{T}{2}$:
\begin{equation}
  \label{eqn:tv-stepu}
  V_1(t, \alpha) = {\beta}\times{H}(t-\alpha) =
  \begin{cases}
    0 & \text{if $t - \alpha  < 0$} \\
    \frac{\beta}{2} & \text{if $t - \alpha  = 0$} \\
    A & \text{if $t - \alpha > 0$} \\
    \end{cases}
  \end{equation}
\begin{equation}
  \label{eqn:tv-stepd}
  V_2(t,\alpha) = \beta - V_1(t, \alpha)
\end{equation}
Eqn.~\eqref{eqn:tv-stepu} models a sudden \emph{increase} in the adversity of the
swarm's operating environment, and Eqn.~\eqref{eqn:tv-stepd} models a sudden
\emph{decrease} in the adversity of the swarm's operating environment. We hypothesize
that GP-DPO swarms will be the most reactive and adaptive of all tested $\kappa$, due
to (1) the relatively high values of $Z(m_i,\kappa)$ previously observed (even
though they are inconsistent across swarm sizes), and (2) their ability to
collectively choose a task less impacted by the applied variance (i.e., utilizing the
central cache).

\begin{figure}[tb]
  \centering
  \includegraphics[width=.9\linewidth]{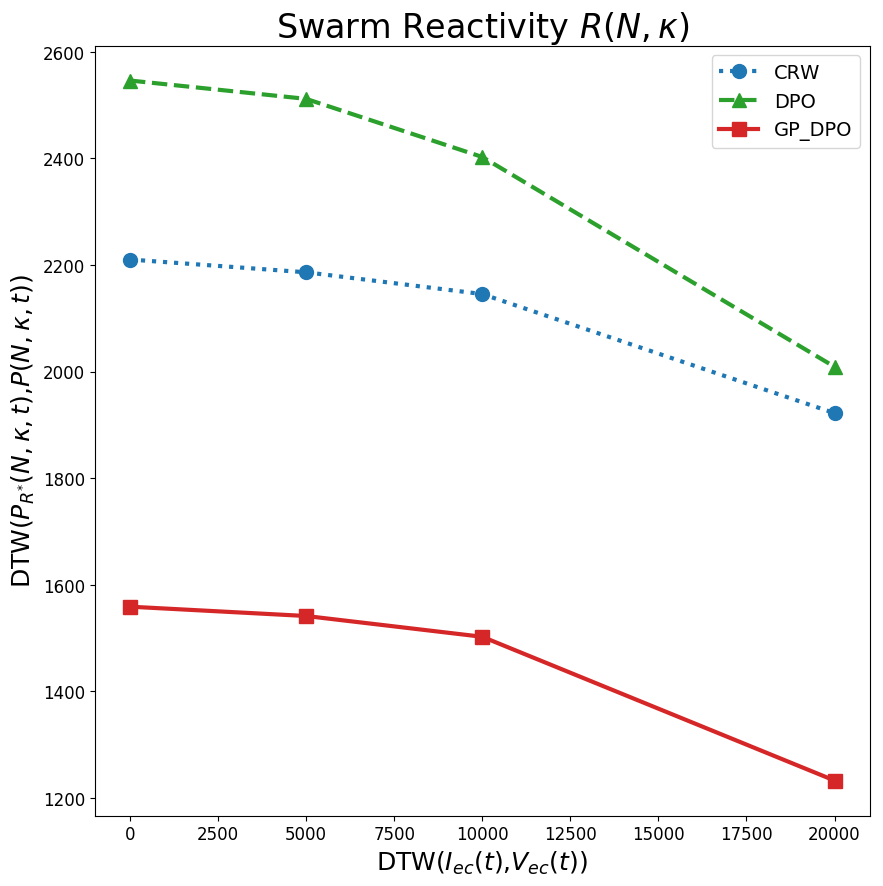}
  \caption{\label{fig:tv-step-reactivity}Swarm reactivity $R(N,\kappa)$ with
    $V_2(t,\alpha)$. Lower Y-values indicate less distance to $P_{R^{*}}(N,\kappa,t)$ and
    therefore greater reactivity.}
\end{figure}

\begin{figure}[tb]
  \centering
  \includegraphics[width=.9\linewidth]{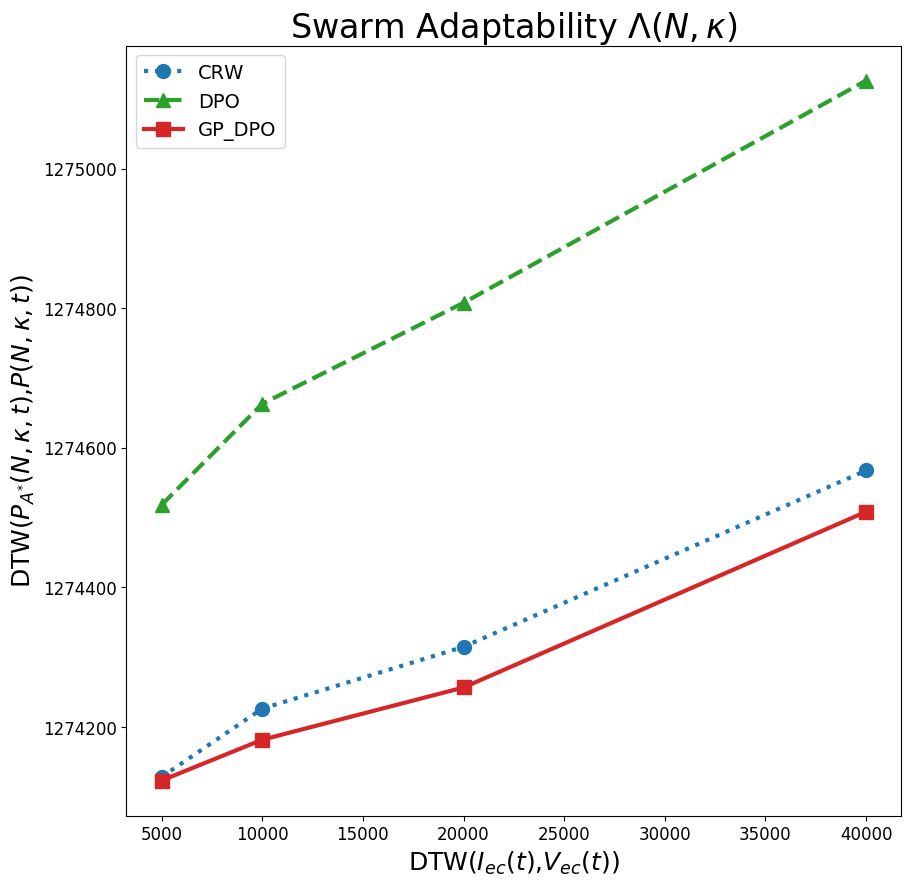}
  \caption{\label{fig:tv-step-adaptability}Swarm adaptability $A(N,\kappa)$ with an applied variance of
    $V_2(t,\alpha)$. Lower Y-values indicate less distance to $P_{A^{*}}(N,\kappa,t)$
    and hence greater adaptability.}
\end{figure}

We see in Fig.~\ref{fig:tv-step-reactivity} and Fig.~\ref{fig:tv-step-adaptability}
that all swarms show increasing levels of reactivity/adaptability with increasing $\beta$
in $V_2(t,\alpha)$, indicating that all $\kappa$ have the ability to proportionally react
and adapt to changing environmental conditions. The plots in
Fig.~\ref{fig:tv-step-reactivity} and Fig.~\ref{fig:tv-step-adaptability} confirm our
hypothesis that the GP-DPO swarms would be the most reactive/adaptive, and
demonstrate for the first time a mathematical framework for equitable comparison of
methods across different scenarios. CRW swarms, while not nearly as reactive, are
also very adaptive, which intuitively makes sense for a biomimetic method.

\section{Discussion}\label{ssec:discussion}

From the progression in Section~\ref{exp-application}, we have been able to use the
insights gained through our proposed measures at each stage of our simulated design
process to build accurate hypotheses about what is likely to occur as we continue to
scale up to meet the demands of our example application. The DPO method provides
almost the same performance with 2,048 robots as the CRW method does with 8,192, but
it is the least reactive and the least adaptive of the chosen methods. Both of these
are important factors in determining method suitability in our target application,
and we therefore select the CRW method with $\sim$8,192 robots to meet our needs. We
did not perform method recommendations for any of the smaller arena and swarm sizes,
though it is clear from the progression above that our proposed metrics can
effectively be used to select the most appropriate method at any scale. While not
rigorous, Section~\ref{exp-application} provides a weakly inductive proof of the
correctness and utility of the proposed metrics within the foraging domain, if not
more broadly.

\subsection{Conclusions and Future Work}\label{sec:conclusion}

We have presented three measurement methods for more precise characterization of
swarm scalability, emergence, and flexibility, in order to provide mathematical tools
to aid in the iterative design process of SR systems. We validated our proposed
measures in the context of a foraging task, and have shown that they provide
intuitive insight into recommending a given method for a given problem size and
operating conditions. A next step in this work would be to develop a method for
quantitative measurement of robustness (e.g., temporally varying swarm sizes and
sensor/actuator noise). In order to facilitate future research and collaboration, the
code used for this research is open source, and can be found at
https://github.com/swarm-robotics/fordyca.

\section*{Acknowledgements}
We gratefully acknowledge Amazon Robotics, the MnDRIVE RSAM initiative at the
University of Minnesota, the Minnesota Supercomputing Institute for their support of
this work, as well as London Lowmanstone's work on developing the initial
experimental framework.


\end{document}